# An Empirical Study of w-Cutset Sampling for Bayesian Networks

**Bozhena Bidyuk**
Information and Computer Science
University Of California Irvine
Irvine, CA 92697-3425
*bbidyuk@ics.uci.edu*

**Rina Dechter**
Information and Computer Science
University Of California Irvine
Irvine, CA 92697-3425
*dechter@ics.uci.edu*

## Abstract

The paper studies empirically the time-space trade-off between sampling and inference in the *cutset sampling* algorithm. The algorithm samples over a subset of nodes in a Bayesian network and applies exact inference over the rest. As the size of the sampling space decreases, requiring less samples for convergence, the time for generating each single sample increases. Algorithm w-cutset sampling selects a sampling set such that the induced-width of the network when the sampling set is observed is bounded by w, thus requiring inference whose complexity is exponentially bounded by w. In this paper, we investigate the performance of w-cutset sampling as a function of w. Our experiments over a range of randomly generated and real benchmarks, demonstrate the power of the cutset sampling idea and in particular show that an optimal balance between inference and sampling benefits substantially from restricting the cutset size, even at the cost of more complex inference.

## 1 INTRODUCTION

Sampling is a common method for approximate inference in Bayesian networks. It is often the only feasible approach (that has some guarantees) when exact inference is impractical due to prohibitive time and memory demands. A significant limitation of all existing sampling schemes, however, is the increase in the statistical variance for high-dimensional spaces. In addition, standard sampling methods fail to converge to the target distribution when the network is not ergodic.

Two well-known variance reduction schemes for sampling methods are blocking([11]) and Rao-Blackwellisation ([6, 3]). Given two strongly correlated variables, we can either sample them simultaneously (blocking) or integrate one of them out (Rao-Blackwellisation). It can be shown that integration is preferred [16]. While Rao-Blackwellisation can be applied in the context of any sampling algorithm, we focus on Gibbs sampling, a member of MCMC sampling methods group, and its Rao-Blackwellised derivative, cutset sampling introduced recently [2].

Given a Bayesian network over the variables $X = \{X_1, ..., X_n\}$, and evidence $e$, Gibbs sampling [7, 8, 17] generates a set of samples $\{x^t\}$ from $P(X|e)$ where each sample $x^t = \{x_1^t, ..., x_n^t\}$ is an instantiation of all the variables in the network. Then we can compute the quantities of interest, such as posterior probabilities, from the samples. Given enough samples, the estimated values are guaranteed to converge to the exact quantities.

The *cutset-sampling* scheme [2] allows sampling over any subset $C$ of the variables $X$, from the distribution $P(C|e)$ at the cost of inference overhead (to compute $P(C|e)$) for each sample generation and for each posterior distribution estimate. By reducing the dimensionality of the sampled space from $X$ to $C$ we are guaranteed to converge over a smaller sample set, yet the cost of generating each sample may increase. The hope is that the increased convergence rate will compensate against the incurred inference overhead. Indeed, the inference overhead created (to compute $P(c^t, e)$) can sometimes be bounded if the structure of the Bayesian network is consulted when selecting the sampled set. This is because exact inference, that can be accomplished by variable elimination of join-tree algorithms ([15, 4, 12]), is controlled by the induced-width of the network whose evidence and sampling nodes are instantiated (the so-called *adjusted induced-width*).

Therefore, when a relatively small cutset induces an inference problem having a bounded adjusted induced-width, the resulting sampling+inference scheme may be more effective than either pure sampling or pure inference. This makes cutset sampling a framework within which we can control the trade-off between sampling and inference and tune its balance to the given Bayesian network.

One point along this trade-off line was already investigated in [2], where we demonstrated empirically that sampling over a cycle-cutset is cost-effective, yielding an order of magnitude improvement over Gibbs sampling.

The contribution of the current paper is in investigating the much broader trade-off range between sampling and inference as a function of w, thus fully evaluating the potential of this scheme. We propose several schemes for finding w-cutset for cutset sampling and evaluate their performance. Using one of the w-cutset selection scheme, we present an empirical study of the cutset sampling scheme over a variety of randomly generated networks, grid structure networks as well as known real-life benchmarks such as CPCS networks. We plot the effect of a gradual change in the cutset-size, as controlled by its adjusted induced width w, on the overall efficiency of w-cutset sampling and show that w-cutset sampling can be highly cost-effective over a range of w's (beyond the case of cycle-cutset). From these experiments we can conclude that cutset-sampling provides a framework within which the user can seek the optimal balance between inference and sampling and tune it to the given network instance.

In Section 2 we provide preliminaries and overview of Gibbs sampling. Section 3 reviews the cutset-sampling scheme. Section 4 discusses statistical bounds of the errors derived from the sampling variance. Section 5 discusses the factors that affect the performance of the w-cutset sampling and defines several greedy w-cutset selection schemes. Section 6 introduces the empirical evaluation which is the primary contribution of this paper. Section 7 provides summary and conclusions.

## 2 BACKGROUND

DEFINITION 2.1 (belief networks) *Let* $X = \{X_1, ..., X_n\}$ *be a set of random variables over multi-valued domains* $D(X_1), ..., D(X_n)$. *A belief network (BN) is a pair* $(G, P)$ *where* $G$ *is a directed acyclic graph on* $X$ *and* $P = \{P(X_i|pa_i)|i = 1, ..., n\}$ *is the set of conditional probability tables (CPTs) associated with each* $X_i$. *A belief network is ergodic if any assignment* $x = \{x_1, ..., x_n\}$ *has non-zero probability, defined by* $P(x_1, ..., x_n) = \Pi_{i=1}^n P(x_i|x_{pa(X_i)})$. *An evidence e is an instantiated subset of variables E.*

DEFINITION 2.2 (induced-width, cycle-cutset) *The width of a node in an ordered undirected graph is the number of the node's neighbors that precede it in the ordering. The width of an ordering d, denoted w(d), is the width over all nodes. The induced width of an ordered graph, w\*(d), is the width of the ordered graph obtained by processing the nodes from last to first. When node X is processed, all its preceding neighbors are connected. A cycle-cutset of an undirected graph is a subset of nodes in the graph that, when removed, results in a graph without*

*cycles. A cycle-cutset of a directed graph (also called loop-cutset) is a subset of nodes that when removed the resulting graph is a polytree.*

### 2.1 Gibbs sampling

Given a Bayesian network $\mathcal{B}$, Gibbs sampling generates a set of samples $x^t = \{x_1^t, x_2^t, ..., x_n^t\}$ where t denotes a sample and $x_i^t$ is the value of $X_i$ in sample t. Given a sample $x^t$, (evidence variables remain *fixed*), a new sample is generated by assigning a new value $x_i^{t+1}$ to each variable $X_i$ from its probability distribution conditioned on the values of the remaining variables:

$$x_i^{t+1} \leftarrow P(x_i|x^t \backslash x_i, e) \qquad (1)$$

Here and elsewhere we will use notation $X \backslash X_i$ to describe a set of variables in X excluding $X_i$.

Once all the samples are generated, we can answer any query using the samples. In particular, posterior marginal belief $P(x_i|e)$ for each variable $X_i$ can be estimated by averaging the conditional marginals:

$$\hat{P}(x_i|e) = \frac{1}{T} \sum_{t=1}^T P(x_i|x^t \backslash x_i) \qquad (2)$$

As the number of samples increases, the probabilities $\hat{P}(x_i|e)$ converge to the exact ones [18, 17] under a few assumptions of the underlying Markov chain. The main requirement for convergence is that the network is ergodic.

In Bayesian networks, the value $P(x_i|x^t\backslash x_i, e)$ depends only on the values of the nodes in the Markov blanket of variable $X_i$ consisting of the node's parents, children, and parents of its children. Therefore [18],

$$P(x_i|x^t \backslash x_i, e) = \alpha F(x_i|x_{pa(X_i)}^t) \prod_{\{j|X_j \in ch_i\}} P(x_j^t|x_{pa_j}^t) \qquad (3)$$

## 3 CUTSET SAMPLING

This section gives a brief overview of the cutset sampling method introduced in [2]. Given a subset of variables $C \subseteq X$, called *cutset* or *sampled* variables $C = \{C_1, C_2, ..., C_k\}$, cutset-sampling generates samples $c^t$, t=1...T, over subspace $C$. Similar to Gibbs sampling, a new sample $c^{t+1}$ is obtained by assigning a new value $c_i^{(t+1)}$ to each variable $C_i \in C$, sampled from the conditional probability distribution:

$$P(c_i|c^t \backslash c_i, e) \qquad (5)$$

Given T samples, the posterior marginals for the sampled variables can be estimated as usual by:

$$P(c_i|e) = \frac{1}{T} \sum_t P(c_i|c^t \backslash c_i, e) \qquad (6)$$

---

**Cutset Sampling**

**Input:** A belief network ($\mathcal{B}$), cutset $C = \{C_1, ..., C_m\}$, evidence $e$.

**Output:** A set of samples $c^t$, $t = 1...T_c$.

1. **Initialize:** Assign random value $c_i^0$ to each $C_i \in C$.

2. **Generate samples:**

For t = 1 to T, generate a new sample $c^t$ as follows:

For i = 1 to m, compute new value $c_i^t$ for variable $C_i$ as follows:

2.1 Using a **join-tree clustering** or variable elimination algorithm $JTC(C_i, (c_{(i)}^t, e))$, compute:

$$P(c_i) = P(c_i | c_{(i)}^t, e) \qquad (4)$$

and Sample a new value $c_i^t$ for $C_i$, from $P(c_i)$ End For i, End For t

---

Figure 1: *Cutset-sampling* Algorithm

For variables that are not in the cutset we compute by exact inference the quantities $P(x_i | c^t, e)$ and then average:

$$P(x_i | e) = \frac{1}{T} \sum_t P(x_i | c^t, e) \qquad (7)$$

The key idea of cutset sampling is that the relevant conditional distributions (eq.(4)) can be computed by exact inference algorithms whose complexity is tied to the network's structure and is improved by conditioning on the observed variables. We use $JTC(X, e)$ as a generic name for a class of variable-elimination or join tree-clustering algorithms that compute the exact posterior beliefs for a variable $X$ given evidence $e$ [15, 4, 12]. The cutset-sampling algorithm is given in Figure 1. It is known that the complexity of $JTC(X, e)$ is time and space exponential in the induced-width of the network's moral graph whose evidence variables $E$ are removed, namely, in the *adjusted induced width*. Therefore, given a parameter $w$, it is easy to describe and control the complexity of cutset-sampling using a notion of $w$-cutset.

**DEFINITION 3.1 (w-cutset)** *Given an undirected graph $G = (V, E)$, if $C$ is a subset of $V$ such that when removed from $G$, the induced width of the resulting graph is less or equal $w$, then $C$ is called a $w$-cutset of $G$ and the adjusted induced width of $G$ relative to $C$ is $w$.*

It is easy to show that:

**THEOREM 3.1 (Complexity of sample generation)** *If $C$ is a $w$-cutset, the complexity of generating a single sample by cutset sampling is $O(|C| \cdot d \cdot n \cdot d^w)$ where $d$ bounds the variables domain size, and $n$ is the number of nodes.*

Computing $P(X_i | e)$ using equation (7) requires computing $P(x_i | c^t, e)$ for each variable which is also exponential in $w$, if $C$ is a $w$-cutset.

**THEOREM 3.2 (Complexity of posterior computation)** *Given a $w$-cutset $C$, the complexity of computing the posterior of all the variables using cutset sampling over $T$ samples is $O(T \cdot |C| \cdot d \cdot n \cdot d^w)$.*

Consequently, for the special case of cycle-cutset, both sampling and estimating the marginal posterior are linear in the size of the network multiplied by the cutset size and the number of samples.

Clearly, we should seek minimal $w$-cutset, those that do not include strict subsets that are also $w$-cutsets. However, determining if a $w$-cutset is minimal is costly, requiring to decide if a subgraph has an induced-width below $w$, which is time exponential in $w$. In our experiments we attempt to generate small $w$-cutset but will not insist on minimality. A cycle-cutset is a $w$-cutset when $w$ is the family size. Yet, it is often not a minimal $w$-cutset.

Values $P(x_i | e)$ obtained by cutset sampling are (1) guaranteed to converge to the exact quantities ([2]) and (2) require fewer samples to converge than full sampling ([9, 3, 16]).

# 4 COMPUTING AN ERROR BOUND

Gibbs sampling provides a simple sampling scheme for Bayesian networks that is guaranteed to converge to the correct posterior distribution in ergodic networks. The drawback of Gibbs sampling compared to many other sampling methods is that it is hard to estimate how many samples are needed to achieve a certain degree of convergence. It is possible to derive bounds on the absolute error based on sample variance for any sampling method if it generates independent samples, for example forward sampling and importance sampling. In Gibbs and other Monte Carlo methods, samples are dependent, and we cannot apply the confidence interval estimate directly.

We can create independent samples restarting the chain after every T samples. Let $\hat{P}_m(x|e)$ be an estimate derived from a single chain $m \in [1, ..., M]$ of length T (meaning containing T samples) as defined in equations (2)-(7). The estimates $\hat{P}_m(x|e)$ are independent random variables. Every time we restart the chain, we randomly assign new values to each sampling variable and this assignment is independent from the results generated in previous chains. If we generate a total of M such chains, the posterior marginals $\hat{P}(x|e)$ will be an average of the M results obtained from each chain:

$$\hat{P}(x|e) = \frac{1}{M} \sum_{m=1}^{M} \hat{P}_m(x|e) \qquad (8)$$

Then, we can use the well-known sample variance estimate for random variables:

$$S^2 = \frac{1}{M-1} \sum_{m=1}^{M} (\hat{P}_m(x|e) - \hat{P}(x|e))^2$$

An equivalent representation for sampling variance is:

$$S^2 = \frac{\sum_{m=1}^{M} \hat{P}_m^2(x|e) - M\hat{P}^2(x|e)}{M-1} \qquad (9)$$

where $S^2$ is easy to compute incrementally storing only the running sums of $\hat{P}_m(x|e)$ and $\hat{P}_m^2(x|e)$. By the Central Limit Theorem, ergodic mean converges to Normal distribution $N(\mu, \sigma)$. Therefore, we can compute confidence interval in the $100(1-\alpha)$ percentile used for random variables with normal distribution for small sampling set sizes ([10]). Namely:

$$P\left[ P(x|e) \in [\hat{P}(x|e) \pm t_{\frac{\alpha}{2},(M-1)} \frac{S}{\sqrt{M}} \right] = 1-\alpha \quad (10)$$

where $t_{\frac{\alpha}{2},(M-1)}$ is a table value from t distribution with $(M-1)$ degrees of freedom. In general, this method may yield confidence interval that is too large to be useful. In the experimental section, we provide results showing 90% confidence interval computed for Gibbs sampler and cutset sampling over 20 Markov chains and analyze the feasibility of using confidence interval as a metrics in evaluating performance of Gibbs and cutset sampling.

## 5   SELECTING W-CUTSET

Selecting an optimal w-cutset for a given w is critical to obtaining good approximations. An optimal w-cutset for cutset sampling can be defined as the w-cutset that allows to obtain the best approximation within the given time period. The quality of the approximation is affected by several factors including cutset side (and associated Rao-Blackwellisation effect), complexity of exact inference (and associated time to generate one new sample), and correlations between sampling variables. Ideally, we want to combine all those factors into a single optimization function, but, in practice, it is hard to measure and predict their joint effects.

Rao-Blackwellisation guarantees improvement with the reduction in sampling set size. In other words, given cutset $C_1 \subset C_2$ and the same number of samples N, then sampling from $C_1$ is preferred. However, at this time, we do not have mathematical tools to quantify the effect of Rao-Blackwellisation and predict the degree of improvement by sampling from $C_1$ compared to $C_2$.

Strong correlations negatively affect the convergence rate of the MCMC methods. It is recommended that if two variables are strongly correlated then either they should be sampled jointly (blocking) or one of them should be summed out (Rao-Blackwellisation) ([16]). In a Bayesian network, we can evaluate the strength of the dependencies between a node and its parents and children based on the entries in the conditional probability tables (CPTs). At the same time, following this criteria, the performance of cutset sampling may deteriorate rapidly as both removing a high-degree node from sampling set or sampling nodes jointly will increase complexity of computation.

The complexity of exact inference depends on cutset size and the induced width w of the network with w-cutset nodes instantiated. Still, the minimal w-cutset is not necessarily most efficient. Given a bound w and w-cutset $C \subset X$, $|X| = N$, in order to generate a new sample $c^t$, we need to compute $|C|$ times the joint probability distribution $P(c^t, e)$ which requires, in the worst case, updating N functions of size $d^w$, where d is the bound on the variable domain size. Thus, the total worst-case computation time to generate a new sample is :

$$O(|C| \cdot N \cdot d^w)$$

However, in practice, the size of the functions in the bucket tree will vary. Given two w-cutsets, $C_1$ and $C_2$ with the same induced width bound w such that $|C_1| < |C_2|$, instantiation of the larger cutset $C_2$ may result in a fewer functions of large size in the bucket tree and allow to generate samples faster.

We have tried several greedy schemes for selecting an optimal w-cutset. Each scheme starts with a set C that contains all nodes in X in topological order except evidence E: $C = X \backslash E$ and then removes nodes from C in some order until no other node removed without violating the maximum induced width bound $w_{max}$.

**Greedy Algorithm (GA)** In the topological order, process nodes in C. Select the next node $C_i$ from C and evaluate the induced width of the bucket tree $w_i$ with nodes $C \backslash C_i$ and E observed. If $w_i \leq w_{max}$, remove node $C_i$ from sampling set: $C = C \backslash C_i$.

**Monotonous Greedy Algorithm (GA)** First, obtain 1-cutset by removing from C (in some order) all such nodes that the adjusted induced width w of the min-fill ordering of nodes $X \backslash C, E$ is bounded by $w = 1$. The 1-cutset becomes a starting sampling set for selection of 2-cutset. We repeat this process selecting a (w+1)-cutset from w-cutset until maximum adjusted induced width $w_{max}$ is reached. Following this scheme, nodes with smaller degrees will be removed from the sampling set first unless they are a part of a large family.

**Heuristic Greedy Algorithm (HG)** In this scheme, we order nodes in C by the size of the Markov blanket (consisting of node's children, parents, and children's parents) which reflects how many conditional probabilities tables sizes would be reduced if a given node was instantiated. Then, the Greedy Algorithm was applied.

The size of the minimal w-cutset found by each algorithm and corresponding time to generate a fixed number of samples (specified in the table after the benchmark's name) are shown in Figure 2 (tested on 1GHz CPU, 128 Mb RAM PC). Overall, the heuristic greedy search usually finds the smallest w-cutset and shows the best time. Surprisingly, a simple greedy search also performs well. Most of the time, it finds w-cutset of the same size as heuristic greedy search or second smallest. The monotonous greedy search consistently yields in performance to the heuristic greedy search (with a few deviations) and performs comparably with simple greedy search. The table in Figure 2 demonstrates many instances where the larger cutset for the same bound w is faster (for example, in cpcs179, GA over $|C| = 9$ is considerably faster than HG over $|C| = 8$) and where the increase in w leads, in fact, to the reduction in computation time, contrary to the worst case analysis expectations (for example, in cpcs360b, HG over 3-cutset of size 21 generates samples faster than HG over 2-cutset of size 27).

| | Alg | w*=2 | | w*=3 | | w*=4 | | w*=5 | | w*=6 | |
|---|---|---|---|---|---|---|---|---|---|---|---|
| | | \|C\| | Time | \|C\| | Time | \|C\| | Time | \|C\| | Time | \|C\| | Time |
| cpcs54 | GA | 24 | 3.73 | 16 | 6.42 | 14 | 11.2 | 10 | 20.8 | 8 | 38 |
| 2000 | MG | 25 | 3.51 | 18 | 5.55 | 15 | 15.9 | 12 | 24.3 | 11 | 30.2 |
| samples | HG | 23 | 3.68 | 15 | 4.67 | 11 | 10.3 | 10 | 15.3 | 8 | 19.1 |
| cpcs179 | GA | 16 | 2.08 | 12 | 5.3 | 9 | 6.04 | 6 | 51 | 5 | 128 |
| 200 | MG | 17 | 2.47 | 12 | 5.22 | 9 | 11.7 | 7 | 33.1 | 4 | 213 |
| samples | HG | 16 | 2.15 | 11 | 2.91 | 8 | 16.6 | 6 | 64 | 5 | 162 |
| cpcs360b | GA | 28 | 4.94 | 22 | 4.12 | 19 | 4.62 | 17 | 6.42 | 17 | 7.74 |
| 100 | MG | 28 | 4.23 | 21 | 4.61 | 19 | 4.58 | 16 | 5.72 | 15 | 9.11 |
| samples | HG | 27 | 4.34 | 21 | 3.73 | 18 | 3.82 | 16 | 5.02 | 16 | 7.01 |
| cpcs422b | GA | 81 | 6.54 | 69 | 5.89 | 60 | 6.59 | 55 | 4.83 | 57 | 10.8 |
| 20 | MG | 83 | 5.61 | 72 | 5.99 | 64 | 4.44 | 58 | 6.75 | 53 | 7.52 |
| samples | HG | 78 | 5.55 | 66 | 5.5 | 57 | 6.16 | 50 | 7.42 | 46 | 8.13 |

Figure 2: Size of the w-cutset and sampling time for a fixed number of samples obtained by different algorithms: GA=greedy algorithm, MG=monotonous greedy algorithm, HG=heuristic greedy algrotihm.

The described schemes for selecting w-cutset are presented primarily to demonstrate the influence of the different factors on the performance of cutset sampling. Those algorithms are not optimal and further investigation is necessary to identify good w-cutset selection methods.

# 6 EXPERIMENTS

## 6.1 Methodology

The primary goal of our empirical study was to investigate performance of w-cutset as a function of w. We compared the performance of Gibbs sampling, w-cutset sampling for different values of w and a special case of w-cutset sampling, cycle-cutset sampling. In all empirical studies, cycle-cutset of the network was found using the mga algorithm ([1]). The w-cutset was selected using monotonous greedy algorithm (MG) defined in Section 5. It is not the most efficient scheme, but its performance is comparable with other w-cutset selection schemes and it guarantees that each $(w + 1)$-cutset is a proper subset of $w$-cutset. Therefore, given the same number of samples, the $(w + 1)$-cutset is predictably better following the Rao-Blackwellisation theory. This property allows us to eliminate some of the uncertainty associated with selecting different sampling sets and focus the empirical study on the trade-offs between cutset size reduction and the associated increase in the complexity of the exact inference as we gradually increase the induced width bound w.

Our benchmarks are several CPCS networks, grid networks, 2-layer networks, random networks, and coding networks. All the sampling algorithms were given a fixed time bound. Small networks, cpcs54 (w*=15) and cpcs179 (w*=8), where exact inference is easy, we allocated 20 seconds. Larger networks were allocated up to 20% of the exact inference time not to exceed 6 minutes.

| | | | #samples | | | | | | |
|---|---|---|---|---|---|---|---|---|---|
| | Gibbs | LC | w*=2 | w*=3 | w*=4 | w*=5 | w*=6 | w*=7 | w*=8 |
| cpcs54 | 1000 | 700 | 700 | 450 | 300 | 200 | 100 | - | - |
| cpcs179 | 130 | 60 | 300 | 80 | 40 | 20 | - | - | - |
| cpcs360b | 500 | 900 | 1000 | 1000 | 900 | 700 | 400 | 300 | - |
| cpcs422b | 10 | 14 | 200 | 200 | 180 | 170 | 140 | 75 | 60 |
| grid | 2000 | 500 | 300 | 260 | 150 | 105 | 60 | 35 | 20 |
| random | 2000 | 1000 | 1400 | 700 | 450 | 300 | 140 | 75 | - |
| 2layer | 200 | 700 | 900 | 320 | 150 | 75 | 40 | - | - |
| coding | 650 | 450 | 800 | 600 | 250 | 150 | 100 | - | - |

| | | | Sampling Set Size | | | | | | |
|---|---|---|---|---|---|---|---|---|---|
| | Gibbs | LC | w*=2 | w*=3 | w*=4 | w*=5 | w*=6 | w*=7 | w*=8 |
| cpcs54 | 51 | 16 | 25 | 18 | 15 | 12 | 11 | - | - |
| cpcs179 | 151 | 8 | 17 | 12 | 9 | 7 | 4 | - | - |
| cpcs360b | 328 | 26 | 28 | 21 | 19 | 16 | 15 | 14 | - |
| cpcs422b | 392 | 42 | 78 | 67 | 59 | 53 | 48 | 43 | 37 |
| grid | 410 | 169 | 163 | 119 | 95 | 75 | 60 | 50 | 13 |
| random | 190 | 30 | 61 | 26 | 25 | 24 | 18 | 17 | - |
| 2layer | 185 | 17 | 22 | 15 | 13 | 12 | 11 | - | - |
| coding | 100 | 26 | 38 | 23 | 18 | 17 | - | - | - |

Figure 3: Markov chain sample count and sampling set size as a function of w.

Each sampling algorithms generated M=20 independent Markov chains of size $T_m = \frac{1}{20}T$ where T is the maximum number of samples that an algorithm could generate within the given time bound. The resulting chain length for each sampling algorithm for different benchmarks as well as sampling set sizes are given in Figure 3. Each chain m produces approximation $\hat{P}_m(x_i|e)$ for the posterior marginals over $T_m$ samples as shown in eq.(8). We obtained a final estimate by averaging over $\hat{P}_m(x_i|e)$ values.

For comparison, we also show the performance of Iterative Belief Propagation (IBP) algorithm on each benchmark after 25 iterations. IBP is an iterative message-passing algorithm that performs exact inference in Bayesian networks without loops ([18]). Applied to Bayesian networks with loops, it computes approximate posterior marginals. The advantage of IBP as an approximate algorithm is that it requires linear space and usually converges fast.

The quality of the approximate posterior marginals is measured by Mean Square Error (MSE) between the exact posterior marginals $P(x_i|e)$ and the approximate posterior marginals $\hat{P}(x_i|e)$:

$$MSE = \frac{1}{\sum_i |D(x_i)|} \sum_i \sum_{D(x_i)} (P(x_i|e) - \hat{P}(x_i|e))^2$$

averaged over the number of instances tried. The exact posterior marginals were obtained via bucket-tree elimination algorithm ([4]). All experiments were performed on 2 GHz CPU 256 Mb RAM PC.

## 6.2 Benchmarks

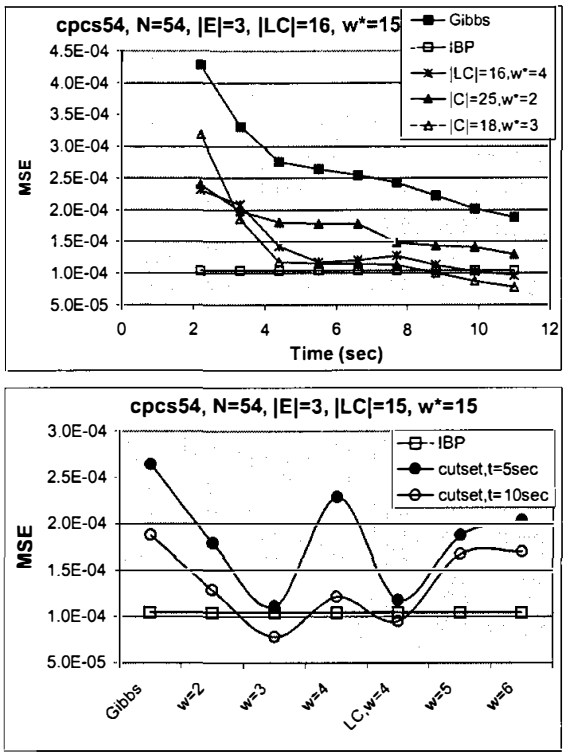

Figure 4: cpcs54, 10 instances, time bound=12 seconds.

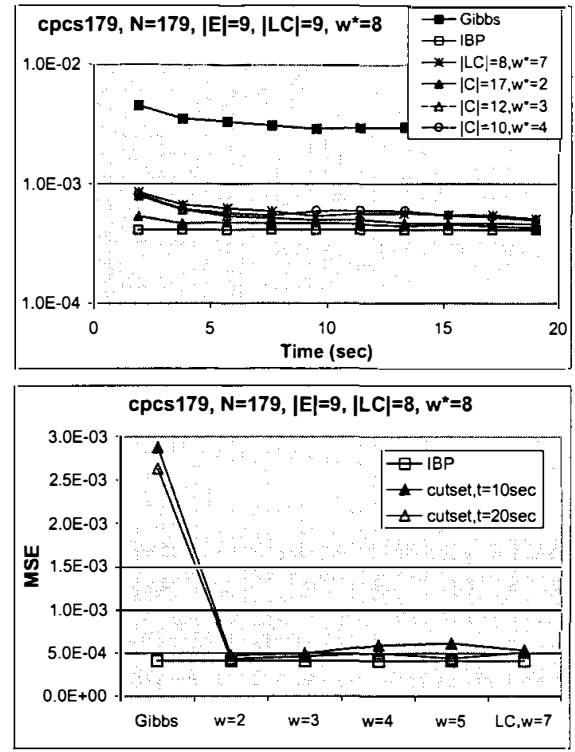

Figure 5: cpcs179, 10 instances, time bound=20 seconds.

**CPCS networks.** CPCS networks are derived from the Computer-based Patient Case Simulation system. The nodes in CPCS networks correspond to diseases and findings and conditional probabilities describe their correlations. For all CPCS networks, we present two charts: one chart demonstrates the convergence over time of Gibbs sampling and w-cutset sampling for several selected w values; the second chart plots the change in the quality of approximation (MSE) as a function of w for two time lines - half of the total sampling time and the total sampling time.

**cpcs54** network consists of $N = 54$ nodes and has a relatively large cycle-cutset of size $|LC| = 15$ (> 25% of the nodes). Its induced width is 15. The performance of Gibbs sampling and cutset sampling is shown in Figure 4. The chart title contains the following notation: $N$ - number of nodes in the network; $|E|$ - average number of evidence nodes; $|LC|$ - size of cycle-cutset; $w*$ - adjusted induced width of the network. The results are averaged over 10 instances with different evidence, $1 - 4$ observed nodes. The first graph, Figure 4, shows the means square error in the posterior marginals as a function of time for Gibbs sampling, cycle-cutset sampling, and w-cutset sampling for w=2 and w=3. The second chart shows accuracy as a function of w. The first point corresponds to Gibbs sampling ($\sim w* = 1$), other points correspond to cutset sampling with different bound w in range from 2 to 6. The cycle-cutset result is embedded with the w-cutset values at

w=4. In cpcs54, the best results are obtained by 3-cutset sampling; the cycle-cutset sampling is second best.

**cpcs179** network consists of N=179 nodes. It has a small cycle-cutset of size $|LC| = 8$ but with a relatively large corresponding adjusted induced width w*=8. The cutset sampling algorithm performance is very similar for all different cutsets (cycle-cutset, 2-,3-,4-,5-cutset) as seen in the accuracy vs. time chart at the top of Figure 5 and at the accuracy vs. w chart at the bottom. The cutset sampling algorithm performed significantly better compared to Gibbs sampler as the network has some non-ergodic properties.

**cpcs360b** is a larger CPCS network with 360 nodes, adjusted induced width of 21, and cycle-cutset $|LC| = 26$. Exact inference on cpcs360b averaged $\sim 30$ min. As we can see from Figure 6), the cycle-cutset sampling and the 2-, 3-, and 4-cutset sampling perform comparatively the same. The top chart in Figure 6 provides a little more insight demonstrating that in the first half of the sampling time period, 3-cutset and 4-cutset perform better. Later, the 2-cutset performs better. This is typical of MCMC methods where the convergence is guaranteed as number of samples approaches infinity but the improvement in behavior may be non-monotonous. All cutset sampling implementations substantially outperform Gibbs sampling taking advantage of both sampling space reduction and greater efficiency in generating samples. All cutset sampling implementations outperform IBP given enough time.

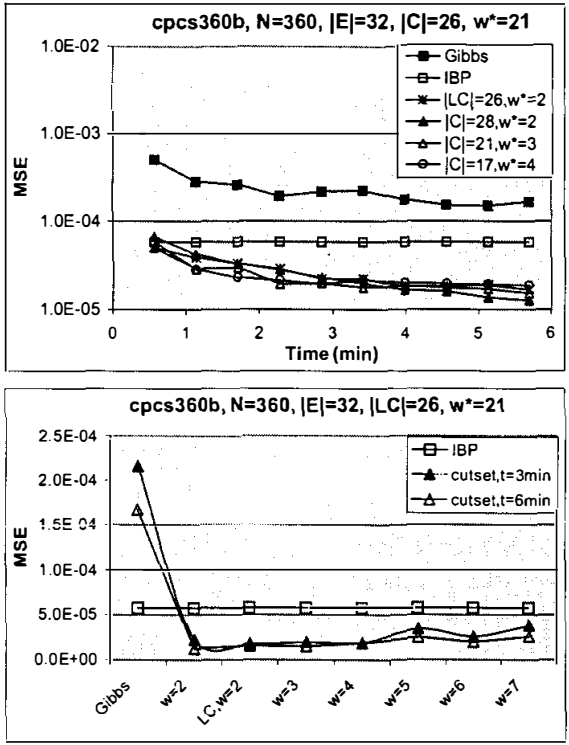

Figure 6: cpcs360b, 10 instances, time bound=6 minutes.

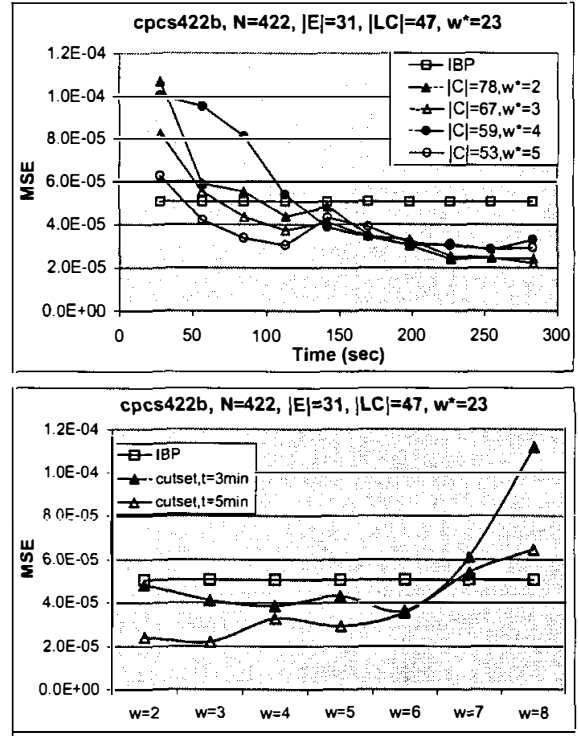

Figure 7: cpcs422b, 3 instances, time bound=5 minutes.

**cpcs422b** is the largest of the CPCS networks with 422 nodes, cycle-cutset size $|LC| = 47$, and induced width $w* = 22$. It also contains several large CPTs so that even when cycle-cutset is instantiated, bucket-tree records the functions of size 14. We observed that w-cutset (but not cycle-cutset!) sampling handled large function sizes efficiently and computed samples an order of magnitude faster than Gibbs sampling or cycle-cutset sampling. In fact, within the 6 minutes time limit given to the sampling algorithm, Gibbs sampling and cycle-cutset sampling could only generate on the order of 10-15 samples each which is statistically insignificant and thus, were left out of the charts. The results for w-cutset sampling are shown in Figure 7. Note that in cpcs422b, the w-cutset was able to take advantage of the network structure to the extent that allowed to increase w efficiently. The bottom chart in Figure 7 shows that w-cutset performed well in range of w=2-8.

**Random networks.** We generated a set of random networks with N=200 binary nodes and r=50 root nodes with uniform priors. Each non-root node $X_i$ was assigned P=3 parents (selected randomly). The conditional probabilities were also chosen randomly, from uniform distribution. Evidence nodes E were selected at random from leaf nodes (nodes without children). The w-cutset sampling significantly improves over Gibbs sampling and IBP reaching optimal perform for w=2-4. In this range, its performance is similar to that of cycle-cutset sampling.

**2-Layer networks.** We generated a set of random 2-layer networks with binary nodes, with r=50 root nodes (with uniform priors), and a total of N=200 nodes (with binary domains). Each non-root node was assigned 3 parents selected at random from root nodes. The CPT values were chosen randomly from uniform distribution. We collected data for 10 instances (Figure 8(a)). As we can see the performance of both Gibbs sampling and IBP is very poor (most likely due to extreme conditional probabilities present in the networks). The w-cutset sampling offers substantial gain in accuracy compared to those algorithms reaching the optimal performance at w=2,3 that is very near the performance of cycle-cutset sampling.

**Grid networks.** The grid networks with 450 nodes (15x30) were the only class of the networks where full Gibbs sampling was able to generate samples faster than cutset sampling and produce comparable estimates. The fastest instance of cutset sampling, in this case cycle-cutset sampling, was 4 times slower compared to Gibbs sampling (see Figure 3). This indicates that networks with regular network structure (that cutset sampling cannot exploit to its advantage) and small CPTs (in a two-dimensional grid network, each node has at most 3 parents) represent a class of networks where Gibbs sampling is a strong player while cutset sampling requires further optimizations. With respect to the accuracy of the estimates, Gibbs sampler, cycle-cutset sampling, and 3-cutset sampling were the best and achieved similar quality results.

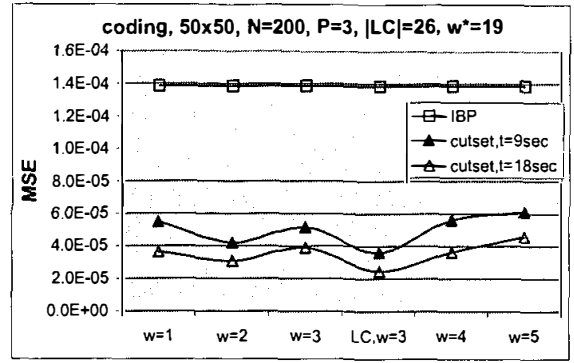

Figure 9: Coding networks, 50 code bits, 50 parity check bits, 100 instances, time bound=6 minutes.

**Coding Networks.** We experimented with coding networks with 200 nodes (50 coding bits, 50 parity check bits). Each parity check bit was assigned three parents from coding bits. The Gallager code parity check matrices were generated using the source code by David MacKay (see http://www.inference.phy.cam.ac.uk/mackay/CodesFiles.html). The results are shown in Figure 9. In this class of networks, the induced speed varied from 18 to 22 making exact inference quite feasible. However, we additionally tested and observed that even a small increase in the network size to 60 code bits, a total of 240 nodes, the induced width exceeds 24 and and exact inference requires considerably longer time while sampling time scales up linearly. We collected results for 10 networks (10 different parity check matrices) with 10 different evidence instantiations (total of 100 instances). In decoding, the Bit Error Rate (BER) is a standard error measure. However, we computed MSE over all unobserved nodes to evaluate the quality of approximate results more precisely. Coding networks are not ergodic due to the deterministic parity check function. As a result, Gibbs sampling did not converge, generated sporadic results, and was left off the charts. At the same time, the subspace of code bits only is ergodic and cutset sampling on a subset of coding bits converges and generates results comparable to those of IBP. Given enough time, cutset sampling can even outperform IBP. The charts in Figure 9 show that cycle-cutset has actually proven to be the best cutset for the coding networks closely followed by 2-cutset sampling.

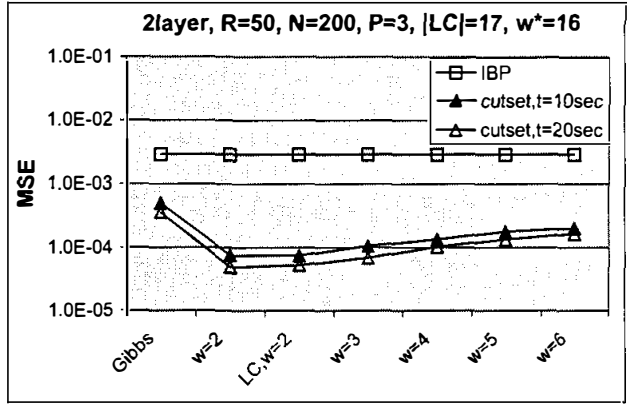

(a) 2layer networks, time bound=25 sec.

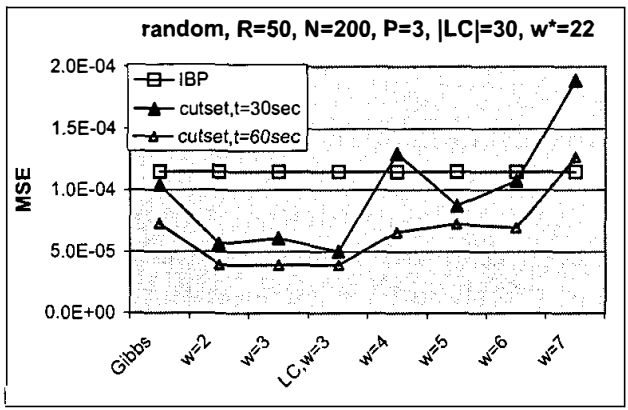

(b) Random networks, time bound=60 seconds.

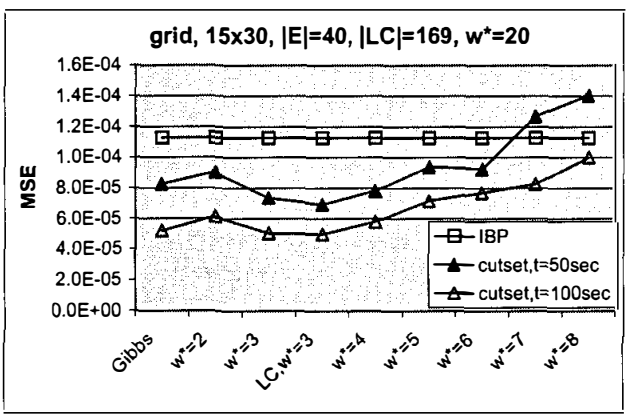

(c) Grid networks, 450 nodes (15x30), time bound=100 seconds.

Figure 8: Randomized networks , 10 instances each.

### 6.3 Estimating Absolute Error

We have computed a confidence interval for estimated posterior marginal $P(x_i|e)$ based on the on the sampling variance of $P_m(x_i|e)$, the estimates produced by independent Markov chains, as described in Section 4. We computed sampling variance $S^2$ from eq.(9) and 90% confidence interval $\Delta_{0.9}(x_i)$ from eq.(10) and averaged over all nodes:

| | Error | Gibbs | LC | w*=2 | w*=3 | w*=4 | w*=5 | w*=6 |
|---|---|---|---|---|---|---|---|---|
| cpcs54 | Exact | 0.00117 | 0.000662 | 0.00076 | 0.000757 | 0.000703 | 0.00068 | 0.001028 |
| | Est | 0.00265 | 0.001308 | 0.00167 | 0.001314 | 0.001547 | 0.00177 | 0.001772 |
| cpcs179 | Exact | 0.02386 | 0.002798 | 0.00204 | 0.002215 | 0.002479 | 0.00198 | - |
| | Est | 0.02277 | 0.002591 | 0.00146 | 0.001814 | 0.002229 | 0.00263 | - |
| cpcs360b | Exact | 0.00109 | 0.000178 | 0.00014 | 0.000189 | 0.000152 | 0.00025 | 0.000372 |
| | Est | 0.00282 | 0.000432 | 0.00049 | 0.000396 | 0.000402 | 0.0006 | 0.000719 |
| cpcs422b | Exact | - | - | 0.00028 | 0.000206 | 0.00032 | 0.00041 | 0.000345 |
| | Est | - | - | 0.00076 | 0.000642 | 0.000575 | 0.00064 | 0.000632 |
| grid 15x30 | Exact | 0.00108 | 0.000992 | 0.00119 | 0.000909 | 0.000986 | 0.00109 | 0.001126 |
| | Est | 0.00248 | 0.002145 | 0.00247 | 0.00205 | 0.002254 | 0.00222 | 0.002386 |
| random N=200 | Exact | 0.00091 | 0.000392 | 0.00039 | 0.000549 | 0.000644 | 0.00062 | 0.001024 |
| | Est | 0.00199 | 0.000799 | 0.00089 | 0.001075 | 0.00124 | 0.00171 | 0.001996 |
| 2layer N=200 | Exact | 0.00436 | 0.000655 | 0.00063 | 0.000815 | 0.001171 | 0.00134 | 0.001969 |
| | Est | 0.00944 | 0.001445 | 0.00144 | 0.001846 | 0.002354 | 0.00302 | 0.003414 |
| coding N=200 | Exact | - | 0.00014 | 0.00019 | 0.000189 | 0.000174 | - | - |
| | Est | - | 0.000296 | 0.00035 | 0.000336 | 0.000356 | - | - |

Figure 10: Average absolute error (exact) and estimated confidence interval (est.) as a function of w.

$$\Delta_{0.9} = \frac{1}{N \sum_i |D(X_i)|} \sum_i \sum_{x_i \in D(X_i)} \Delta_{0.9}(x_i)$$

As noted earlier, estimated confidence interval can be too large to be practical. Thus, we compared $\Delta_{0.9}$ with exact average absolute error $\Delta$:

$$\Delta = \frac{1}{N \sum_i |D(X_i)|} \sum_i \sum_{x_i \in D(X_i)} |\hat{P}(x_i|e) - P(x_i|e))$$

The objective of this study was to observe whether computed confidence interval $\Delta_{0.9}$ (estimated absolute error) reflects true absolute error $\Delta$, that is $\Delta < \Delta_{0.9}$ ? and how big the estimate is compared to the true error since large confidence interval is not very informative.

Figure 10 shows the average confidence interval and average absolute error for a set of benchmarks described earlier. As we can see, for all of the networks except cpcs179, we observe that $\Delta < \Delta_{0.9}$. In cpcs179, the estimated bound is sometimes a little smaller than average error which maybe attributed to non-ergodic properties of cpcs179 (even though the variance is small, the algorithm did not explore the sampling space properly) and stochastic nature of the estimate. Also, we observe that in most cases the estimated confidence interval is no more than 2-3 times the size of average error. Thus, we can conclude that confidence interval estimate could be used as a criteria reflecting the quality of the posterior marginal estimate by the sampling algorithm where the comparison against exact posterior marginals is not possible.

### 6.4　Summary

The empirical evaluation of the performance of w-cutset sampling shows that with the exception of grid networks, w-cutset sampling always outperforms Gibbs sampling (in some instances, generating samples even faster than Gibbs), outperforms cycle-cutset sampling for some w values, and offers a considerable improvement over IBP on several networks. We have discovered a class of networks where we can recommend with a degree of certainty Gibbs or cutset sampling. Gibbs appears to have an advantage in the networks with regular structure and small probability tables (such as grid networks). Cutset sampling performs well when it can take advantage of the underlying network structure and the network contains a few large probability tables (as in cpcs422b).

Overall, we can conclude that there exists a range of w values where w-cutset sampling achieves an optimal performance. In many instances, increasing w, up to some threshold value, compensates for the incurred overhead in exact inference due to variance reduction and, in some instances, also due to increased speed of generating samples (discussed in Section 5). The performance of w-cutset begins to deteriorate when increase in w results only in a small reduction of sampling set size. An example is cpcs360b network where starting with w=5, increasing w by 1 results in the reduction of sampling set only by 1 node (shown in in Figure 2) and leads to the reduction in the size of sampling chain (shown in in Figure 3).

Further, our empirical study conclusively indicates that in most cases cycle-cutset makes a good sampling set. It is not always the best, but usually it is comparable to the best w-cutset results for some w and maybe practical when we do not want to invest time in finding an optimal w-cutset.

Finally, the comparison of estimated confidence interval for approximate posterior marginals to the exact absolute error indicates that confidence interval accurately reflects the absolute error (most of the time, within the factor of 2-3 and, in case of cutset sampling, never exceeding the range of $[0, 0.003]$) and could be used as a measure of sampling algorithm performance.

## 7　RELATED WORK AND CONCLUSIONS

In this paper, we investigated the performance of w-cutset sampling, combining sampling and exact inference, as a function of the adjusted induce width parameter w that controls the complexity of the exact inference. Our experiments over a range of randomly generated and real benchmarks, demonstrate the power of the cutset sampling idea and in particular show that an optimal balance between inference and sampling benefits substantially from restricting the cutset size, even at the cost of more complex inference. Thus, user can control the trade-off between sampling and inference by examining the w-cutset sampling for different w values. We have proposed in this paper several greedy schemes for finding a w-cutset for cutset sampling. In future study, we will continue to investigate the trade-off between cutset size and their w values based on the network structure in order to determine ahead of time a good w-cutset for the scheme. Alternatively, we can propose to the user a bag-of-algorithms for selecting a w-cutset and allow the user to determine the most efficient cutset empirically.

Previously, sampling from a subset of variables was successfully applied to a particle factoring using importance sampling for Dynamic Bayesian networks (DBNs) [5]. However, in [5], the authors narrowly targeted the class of DBN networks where each time slice contains nodes independent from previous time slice and, therefore, are easy to sum over. The cutset sampling, first defined in [2], applies the Rao-Blackwellisation scheme in a more general setting of any Bayesian network.

A different combination of sampling and exact inference for join trees was described in [14] and [13]. Each of those approaches proposes a scheme for sampling locally within a cluster and then combining those distributed results efficiently (introducing additional errors in the result). In [11], exact inference was used in combination with blocking Gibbs sampling. The cutset sampling is fundamentally different from all of those approaches in that it sums out variables instead of blocking or localizing the sampling.

### Acknowledgements

This work was supported in part by NSF grant IIS-0086529 and MURI ONR award N00014-00-1-0617.